\documentclass[journal]{vgtc}                
\usepackage[mathlines,switch]{lineno}

\ifpdf
  \pdfoutput=1\relax                   
  \usepackage{graphicx}                
  \DeclareGraphicsExtensions{.pdf,.png,.jpg,.jpeg} 
\else
  \usepackage{graphicx}                
  \DeclareGraphicsExtensions{.eps}     
\fi%

\graphicspath{{figures/}{pictures/}{images/}{./}} 

\usepackage{microtype}                 
\PassOptionsToPackage{warn}{textcomp}  
\usepackage{textcomp}                  
\usepackage{mathptmx}                  
\usepackage{times}                     
\usepackage{cite}                      
\usepackage{tabu}                      
\usepackage{booktabs}       
\usepackage[table, svgnames, dvipsnames]{xcolor}
\usepackage{scalerel}
\usepackage{changes}
\usepackage{mathtools}
\usepackage{enumerate} 

\newcounter{lenumerate}

\renewenvironment{itemize}{\begin{list} {\labelitemi}
{\setlength{\parsep}{0.1ex} \setlength{\topsep}{0.6ex}
\setlength{\partopsep}{0.1ex } \setlength{\itemsep}{0.2ex}
\setlength{\leftmargin}{2ex} }} {\end{list}}

\onlineid{1063}

\vgtccategory{Research}

\title{Interactive Visual Study of Multiple Attributes Learning Model of X-Ray Scattering Images}

\author{Xinyi Huang, Suphanut Jamonnak, Ye Zhao, Boyu Wang, Minh Hoai, Kevin Yager, Wei Xu}
\authorfooter{
\item X. Huang, S. Jamonnak, Y. Zhao are with Kent State University (Email: xhuang, sjamonna, zhao@cs.kent.edu). 
\item B. Wang, M. Hoai are with Stony Brook University (Email: boywang, minhhoai@cs.stonybrook.edu). 
\item K. Yager, W. Xu are with Brookhaven National Laboratory (Email: kyager, xuw@bnl.gov).
}

\abstract{
   Existing interactive visualization tools for deep learning are mostly applied to the training, debugging, and refinement of neural network models working on natural images. However, visual analytics tools are lacking for the specific application of x-ray image classification with multiple structural attributes. In this paper, we present an interactive system for domain scientists to visually study the multiple attributes learning models applied to x-ray scattering images. It allows domain scientists to interactively explore this important type of scientific images in embedded spaces that are defined on the model prediction output, the actual labels, and the discovered feature space of neural networks. Users are allowed to flexibly select instance images, their clusters, and compare them regarding the specified visual representation of attributes. The exploration is guided by the manifestation of model performance related to mutual relationships among attributes, which often affect the learning accuracy and effectiveness. The system thus supports domain scientists to improve the training dataset and model, find questionable attributes labels, and identify outlier images or spurious data clusters. Case studies and scientists feedback demonstrate its functionalities and usefulness.
} 

\teaser{
  \centering
  \includegraphics[width=\textwidth]{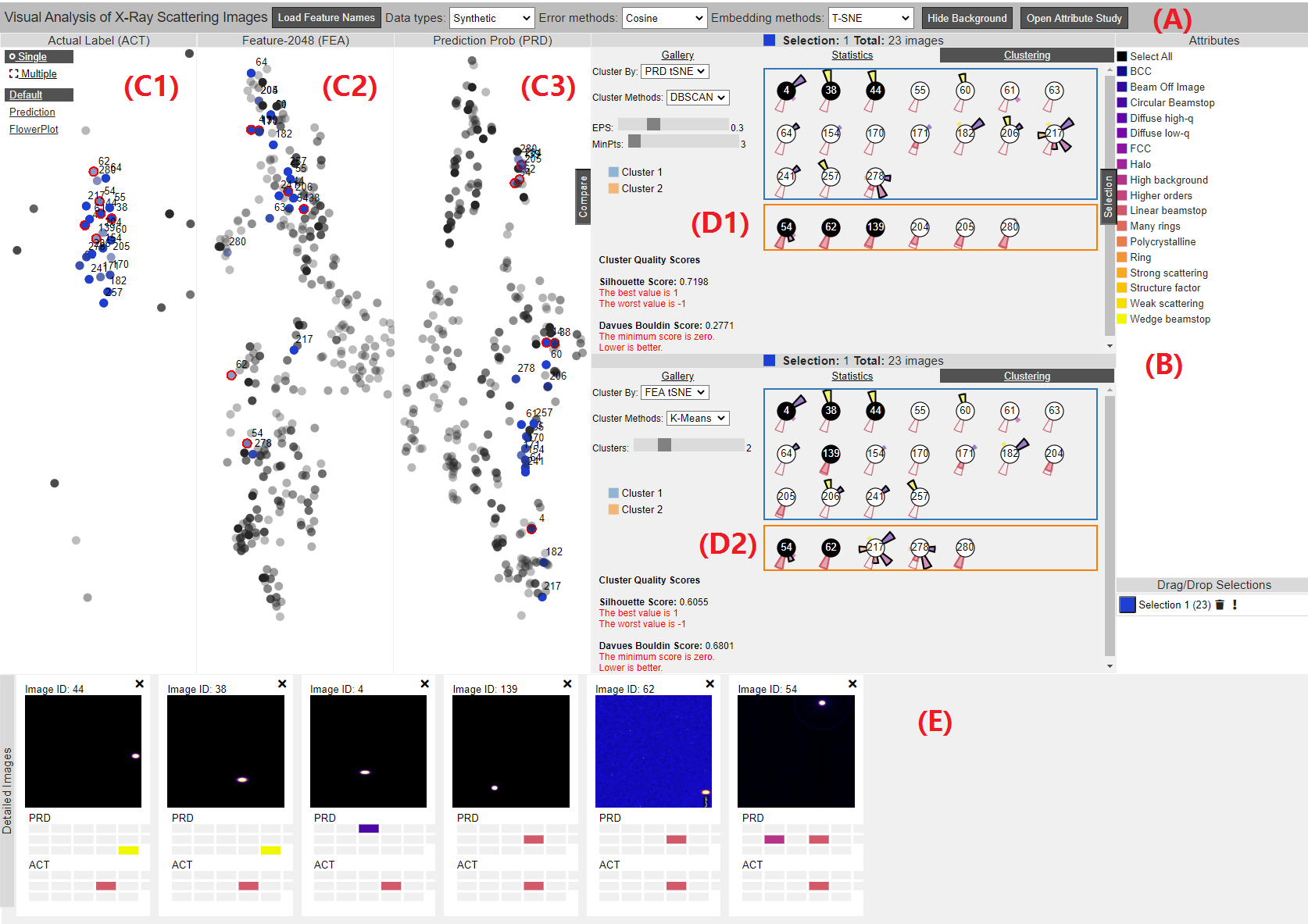}
    \caption{ System interface: (A) Control panel. (B) Attribute panel. (C1) Embedded view in ACT (Actual attribute vector) space with a highlighted image group in red. (C2) Embedded view in FEA (Feature vector) space. (C3) Embedded view in PRD (Prediction vector) space. (D1) Group panel 1:  a cluster view showing selected images as attribute flowers in two spatial clusters of PRD (C3). (D2) Group panel 2: a cluster view showing the images in two spatial clusters of FEA (C2). (E) Detail image view with ACT/PRD values. }\vspace{-1pt} 
    \label{fig_interface} 
}

 \nocopyrightspace

\vgtcinsertpkg

\begin{document}



\firstsection{Introduction}
\maketitle

X-ray scattering helps scientists discover molecular and nano-level physical structures of materials such as nano-particles, protein, lithographic gratings, polymer films, and so on. X-ray beam hits on a sample material, and the scattered x-ray diffraction patterns are collected by the detector which are eventually presented in the x-ray scattering images. This technique is widely used in biomedical, material, and physical applications by analyzing structural patterns in the x-ray scattering images \cite{Yager2014}. In general cases, human experimenters must apply their domain expertise to interpret the features in the image (such as rings, diffraction spots, or diffuse scattering) in order to understand the image and select the most appropriate follow-up analysis. However, the scale of an image dataset (an x-ray equipment generates up to 1 million images per day) often imposes a heavy burden in image screening and understanding.

Recently, deep learning models are employed in classifying and annotating multiple image attributes from experimental or synthetic images, which were shown to outperform previously published methods \cite{Wang2017,Guan2018}. As most deep learning paradigms, these methods are not easily understood by material, physical, and biomedical scientists. The lack of proper explanations and absence of control of the decisions would make the models less trustworthy. While considerable effort has been made to make deep learning interpretable and controllable by humans \cite{Choo2018}, the existing techniques are not specifically designed for the scientific image classification models of x-ray scattering images, which requires extra consideration in finding: 
\begin{itemize}
    \item How the learning models perform for a diverse set of overlapped image attributes with high variation?
    \item How the co-existence of attributes in x-ray images may affect the classification results? 
\end{itemize}
Unfortunately, few existing visualization tools \cite{Choo2018} are devoted to visually analyzing the learned results of multiple attributes, objects, or segments with these models.

In response, we develop a visual analysis system for users to interactively study the model predictions with respect to the multiple structural attributes of x-ray scattering images. The system has several features:
\begin{itemize}
    \item Image instances are projected with t-SNE and visualized in three vector spaces: actual labeling space from domain scientists, feature space extracted by a residual network, and prediction space of the model output. Users can explore the instances in these spaces interchangeably for visual comparison of different groups, outlier detection, and drill-down study of images.
    \item Users can select any image group and then observe their visual features and study the attribute detection performance. In particular, the distributions of these images in different spaces are easily discovered by user controlled spatial clustering in the embedded spaces.  
    \item An attribute-flower visualization is designed to represent an image to manifest the attribute recognition results. Compared with the ground truth labels, it depicts false positive (FP), false negative (FN), true negative (TN), and true positive (TP) predictions of multiple attributes.
    \item The learning outcome of the images with different groups of co-existed attributes is visualized, which provides visual cues of model performance with respect to attributes relationships.
    \item The visual interface integrates multiple coordinated views and includes easy user interactions that facilitate iterative exploration and comparison.
\end{itemize}
The system alleviates domain scientists' efforts in understanding the performance of deep learning models for x-ray scattering images. They can identify outlier images, find spurious data clusters, understand the impacts of multiple attributes, and from this improve the training data or the learned model. Several case studies show the utility and effectiveness of the system. Domain scientists provide positive feedback about the usefulness of this visual interaction tool.
\section{Related Work}
It is of great importance to analyze x-ray scattering images by recognizing structural attributes such as ring, halo, diffuse scattering, and so on \cite{Kiapour2014}. Recently, deep learning models are employed for the x-ray scattering data \cite{Sullivan2019, Park2017,Liu2019}. Wang et al. \cite{Wang2017} apply Convolutional Neural Networks (CNN) over both experimental and synthetic images to detect important attributes. Guan et al. \cite{Guan2018} further develop a DVFB-CNN model (Double-View Fourier-Bessel CNN)  which combines Fourier-Bessel transform (FBT) with a CNN model. With the improvement of the performance by different designs of models, the application of deep learning in this domain would make the attribute annotation convenient for the domain experts. 

It is also important to open the black-box by model interpretation to build confidence for domain experts to apply their models.
Choo et al. \cite{Choo2018} categorize explainable deep learning into three major directions: model understanding, model debugging, and model refinement. Computational approaches discover important scores of the input features contributing to the prediction results. Perturbation methods \cite{Alvarez-Melis2017}, saliency-based methods \cite{Selvaraju2016}, LIME \cite{Ribeiro2016}, and influence functions \cite{Koh2017} are proposed for the purposes. In particular, class activation map (cam) \cite{ZhouZ2016} and grad-cam method \cite{Selvaraju2016, Selvaraju2017} can detect pixel areas of one individual image which contributes to a corresponding class prediction. They find the important neurons in a hidden layer and explain them by highlighting direct visible and understandable features in every single image. However, these methods are not designed to understand the model performance of a large set of images with multiple attributes. Our system does not focus on mathematical algorithms that find features in an image linking to classification results and then visualizing them such as by image heatmaps. Instead, our system is built up on studying the distributions of many images in different data spaces.

Interactive visualization tools are developed in providing in-depth understanding of how deep learning models work. Tools such as Tensorflow  Playground\cite{Smilkov2017DirectManipulationVO}, Tensor Board \cite{Wongsuphasawat2018}, and ConvNetJS \cite{ConvNetJS} allow users to visualize and interact with the activation maps and network structures, together with line graphs and histograms of characteristic statistics. DeepVis \cite{Yosinski2015} shows that optimizing synthetic images with better natural image priors produces more recognizable visualizations. CNNVis \cite{Liu2017} system helps designers in their understanding and diagnosis of CNNs by exploring the learned representations in the graph layout of the neural networks. ActiVis \cite{Kahng2018} integrates an embedding view with multiple coordinated views for visual model exploration. Embedding Projector \cite{EmbeddingProjector} visualizes input images in a 2D or 3D embedding space (by PCA or t-SNE), to reveal the relationship among these instances. For studying the training process, Deep View \cite{Zhong2017} presents a level-of-detail framework that measures the evolution of the deep neural network both on a local and on a global scale. Recently, visualization tools \cite{Wang2018a,Liu2018,Kahng2019} have been developed for studying deep generative models (e.g. GAN) working on benchmark natural images. These methods have not been specifically designed for x-ray scattering images.

Our approach allows users to analyze the results of physical attributes recognition in special sets of scientific image data, where multi-attributes may have inherent correlations and co-exist in one x-ray image. And these relationships among attributes may play an important role and affect the performance of the trained model predictions for each image. Therefore, it is different from those existing tools designed for explaining disjoint multi-attributes classification \cite{Pezzotti2018,Rauber2017,Rauber2016,EmbeddingProjector,Maaten2008,Zhong2017}. The scientific attributes have large structural variations, for example, rings can be very large circular structures and can also be very small and indiscernible by eyes \cite{Zhao2019}. The attributes are also correlated in many cases (see Sec. \ref{sec:network}). Based on these features, existing tools cannot be directly applied. Effective visualization tools are needed to discover the relationship among the scientific structural objects on scientific images and the performance of deep learning models. Our system is different from most existing methods, in which image, audio, and natural language datasets are projected and visualized by linking the final decision with the origin images/text data elements. Our visualization techniques focusing on multiple structures on the images may also be extended to multiple object detection and segmentation models, such as (fast-, faster-) R-CNN \cite{Zhao2019}, YOLO \cite{Redmon2017}, and SegNet \cite{Badrinarayanan2017}. 

\section{Neural Network Learning of Multiple Structural Attributes}

In this section, we introduce x-ray images with multiple attributes and the deep learning model.

\subsection{X-Ray Scattering Images and Attributes} \label{sec:network}


\begin{figure}[t]\centering
 \includegraphics[width=0.99\columnwidth, height = 0.85\columnwidth ]{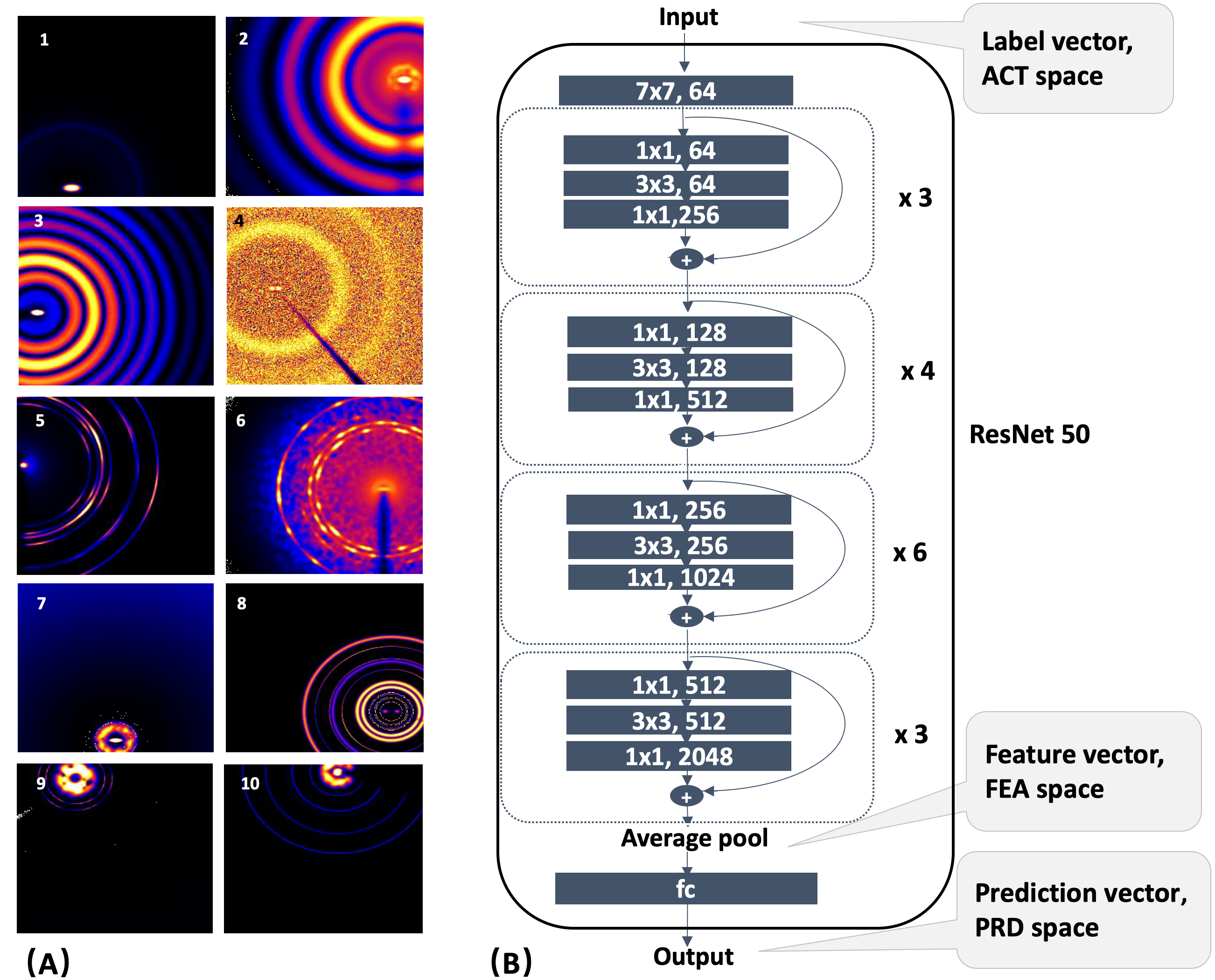}\vspace{-10pt}
    \caption{ (A) X-ray scattering images (1-10) including 17 attributes (see details in Appendix). (B) ResNet50 architecture \cite{He2016} and three data spaces. }\vspace{-15pt}
    \label{fig:resnet}  
\end{figure}

Automatic attribute recognition in x-ray scattering image data is a challenging problem due to the high variation in the same classes. The same structural attributes (patterns) can be of great variety in their appearances. On the other hand, a diverse set of characteristic attributes may co-occur and overlap in the scattering images. The attributes represent the following information about the material being probed in scientific experiments:
\begin{itemize}
    \item Experimental conditions, such as the type of beamstop used to block the x-ray beam (e.g., linear or circular beamstop) or the detector position (e.g., beam off image);
    \item Scattering patterns either holistically (e.g, structure factor, high background, strong/weak scattering) or based on visible features (e.g., ring, many rings, halo, high order, diffuse high/low-q);
    \item Material structures implied by the scattering, such as BCC (body-centered cubic), FCC (face-centered cubic), and polycrystalline;
\end{itemize}
Fig. \ref{fig:resnet} shows nine images including 17 structural attributes (see Appendix for details). The attributes have relations to each other. Some may be highly correlated while others may be mutually exclusive. In some cases, these relationships are inherent to the definitions, while in other cases they may only be ascertained by exploring correlations in real experimental datasets. It is thus important to evaluate how the trained models have learned these attributes.

In this work, we utilize an open x-ray scattering dataset \cite{Yager2017}, and an updated ResNet model \cite{Wang2017} that was designed for multiple attributes classification of the dataset. About 1,000 x-ray scattering images were employed in our visualization system. They include different types of images including semiconductors, nano-particles, polymer, lithographic gratings, and so on. The attributes in these images are either labeled by domain experts or synthetically generated by a simulation algorithm \cite{Wang2017}. Each image thus has an actual attribute vector (\textbf{ACT vector}) consisting of 17 Boolean (0 or 1) values to show if the image has a number of the 17 attributes.

\subsection{ResNet Deep Learning Model and Data Spaces}
The original 50-layer Residual Network (ResNet) proposed in \cite{He2016} is designed for the classification of mutually exclusive image attributes, where the softmax cross-entropy is used as the loss function. For multiple attributes classification of x-ray images, its loss function is modified \cite{Wang2017} by computing the sigmoid cross-entropy for each attribute and then defining the final loss function as the sum of the losses incurred by each attribute. This modified ResNet model \cite{Wang2017} was trained for 17 attributes by more than 100,000 x-ray images. Our system is built up on this model.  

Fig. \ref{fig:resnet} shows the abstract architecture of this model. It has the first 49 hidden convolutional layers for the feature extraction and 1 fully connected layer for classification. A feature vector (\textbf{FEA vector}) with 2048 dimensions is learned by the model for each image. This FEA vector is obtained at the end of the feature extraction phase after average pooling. Then, a fully connected layer is used to generate the final output, i.e. a 17-dimensional prediction vector (\textbf{PRD vector}) representing the predicted structural attributes. The deep learning model reported a mean average precision (mAP) is about 77\% \cite{Wang2017}. 

From the model, the x-ray images are represented by vectors in three different spaces (i.e., ACT, FEA, and PRD). The ACT vector has 17 dimensions with each being the Boolean label corresponding to one attribute. The PRD vector with the same size 17 is obtained from the output of the trained ResNet model with fully connected layers for classification \cite{Wang2017}. Each PRD element is the prediction value (0.0 to 1.0) for each attribute with a cut-off value of 0.5 for the final decision. The FEA vector with the size of 2048 dimensions contains the activation values of the last feature extraction layer in ResNet \cite{He2016}. The characteristics of
an image are supposed to be extracted and representative by its FEA vector, though it is not directly interpretive.

Studying and comparing the x-ray scattering images in these spaces can reflect the performance of the ResNet model. For example, a group of images has similar FEA vectors means the model extracts similar feature elements. If they also have similar PRD vectors, it shows that the model well utilizes the extracted features in the classification phase. On the other hand, these images may not be close in ACT space, which indicates that the actual labels do not agree on the classification results. In such cases, users can study these images to identify labeling errors or find model design issues. Therefore, our system is designed to help users interactively study the x-ray images in these three spaces simultaneously.

\subsection{Model Performance Measures}
Model performance metrics help users understand and study the model output and performance. They will play the role of visual cues for effective interactive exploration. For a set of x-ray scattering images, the prediction output of the learning model generates the standard classification evaluation including FP, TP, FN, and TN for each specific attribute, as well as the metrics including the accuracy, precision (recall) and the F1 score. 





\subsection{Co-existence Relationship of Attributes}  \label{sec:coexistence}

Having multiple attributes in the same images is a unique feature of x-ray scattering images. It is of interest and importance for scientists to discover the impact of attribute relations for model performance. 

\subsubsection{Pairwise co-existence} 
For two attributes, we employ the measures of Pearson correlation, mutual information, and conditional entropy. Pearson correlation coefficient can be used to measure the linear dependence of two attributes. For two vectors $x$ and $y$ representing distributions of two attributes in a set of images, the correlation $r_{\scaleto{\mathbf{x}\mathbf{y}}{3pt}}=0$  indicates that the two attributes have no linear relationship, while $r_{\scaleto{\mathbf{x}\mathbf{y}}{3pt}}=1 (-1)$ shows perfect positive (negative) linear dependence. In addition, mutual information can be used to quantify the mutual dependence of two attributes. It measures the reduction of uncertainty of one attribute due to the knowledge of another attribute. Moreover, conditional entropy can also measure the uncertainty of one attribute, under the conditional state of another attribute. It can reflect the weighted average uncertainty of one attribute, given one known attribute. Please see Sec. \ref{sec:attributes} for their usage.




\subsubsection{Multivariate co-existence}

A variety of metrics is presented to measure the multivariate relationship with more than two attributes \cite{Timme2014}. For example, \textit{total correlation} measures relationships with more than two variables and \textit{interaction information} extends the concept of  mutual information to many variables.  However, based on our experiments with domain scientists, they seldomly use such multivariate measures in their work, and these metrics are hard to understand when analyzing x-ray image results. Therefore, we do not employ them but instead enumerate all real combinations of multiple attributes within the image dataset, for example, any pair of two attributes, and 3-tuples of three attributes, etc. For each combination, the prediction accuracy is represented by finding whether all the involved attributes are correctly detected. By visualizing this information, users can find which set of co-existed attributes is interesting for further study. A large number of attributes (more than 8) do not appear together in one image so that this approach is valid.

\section{Design of the Visualization System}
The visualization system is designed for domain scientists to analyze the model that classifies x-ray images of multiple attributes. Its interface and functions are designed based on a set of analysis tasks.

\subsection{Analysis Task Characterization} 
Domain scientists currently often use the statistical metrics such as precision, recall, and accuracy for learning model evaluation of x-ray scattering images. However, this only presents a general class-wise evaluation rather than an in-depth investigation connecting the groups of images and their various structural patterns. Visual analytics tools can play a vital role in model understanding through interactive visualizations directly over images and attributes. We scheduled several interview meetings with domain scientists who wanted to understand the modified ResNet model behavior over the x-ray scattering image datasets. In these meetings, we conducted a requirement analysis by discussing the topics and their preferred functions in visually analyzing the trained model and datasets. A set of visual analysis tasks were identified as: 

    \noindent \textbf{T1. Analysis in Model Spaces:} Investigate scientific images within the spaces of ACT, PRD, and FEA, for users to understand how the images are modeled by the ResNet in the feature space and then classified by fully connected layers in the prediction space, with respect to the real labels. Users can study ResNet model performance by comparing the distributions of images after feature extraction (FEA), after classification (PRD), and with actual labels (ACT). This study needs to be performed in an exploratory process. Therefore, it is important to visualize the images in the three spaces at the same time.

    \noindent \textbf{T2. Analysis with Group Behaviors:} Select and examine specific groups of images in the ACT, PRD, and FEA spaces, in order to find important clusters and outliers with respect to the learning model. 
    
    \noindent \textbf{T3. Analysis with Image Attributes:} Identify important image instances with the performance metrics of individual attributes and co-existent attributes to perform the first two tasks. 
    
    \noindent \textbf{T4. Analysis with Comparisons:} Compare individual images and image clusters for the model prediction performance.
    

\subsection{Visualization Design and Interface Overview}  \label{sec:design}
With respect to these tasks, we design the visualization system by integrating a set of visual interaction functions  including:
\begin{itemize}   
\item {\bf For T1: Coordinated Visualization in Embedded Spaces:} Images are visualized in the 2D canvases of ACT, PRD and FEA spaces, respectively. The goal is to allow users to (1) easily observe many x-ray images and their relationships in these spaces simultaneously, and (2) interactively select, compare, and study images of interest. Therefore, the 2048-dimensional FEA space and 17-dimensional ACT and PRD spaces are projected to the embedded 2D spaces to fulfill the goal. In our system, we have included two commonly used dimension reduction (DR) algorithms, t-SNE and PCA, for deep learning visualization. Other DR methods may further be added. 

\item {\bf For T2: Image Group Selection and Visualization:} 
Within the embedded spaces, users are enabled to flexibly select images into groups at each embedded space by lasso tools. Then the selected images in each group are visually highlighted in other spaces. This function is very important for users to freely explore images of interest and conduct comparative analysis among the three spaces. The grouped images are also visualized by a statistic view of image metrics and an image gallery view. They can be further clustered for drill-down study and comparison. 

\item {\bf For T3: Attribute Co-existence Visualization:} The model performance with the relations of co-existing attributes is visualized in an interactive view. Users can then define image groups based on the visual cues of model performance. 

\item {\bf For T4. Group Comparative View and Image Comparison:} The selected groups can be easily investigated and compared with group panels for detailed views. Through interactive selection over all the above visualizations, users can also open multiple images to compare their details of raw data and model predictions.
\end{itemize}

Fig. \ref{fig_interface} shows the visualization system interface. It displays the scientific images in the coordinated views (C1-3) of three different embedded spaces (ACT, FEA, PRD). Users can interactively select (with zooming, panning, and lasso-selection) image groups in either embedded view, which is highlighted to show their distributions in the other views. Images in the embedded space are shown as dots whose transparency indicates the model predicted errors. One limitation is that the transparent dots may overlap and lead to a misleading ``artificial'' transparency value (See Sec. \ref{sec:limitation}). Users can filter the visualizations with single or multiple attributes (B). Moreover, the co-existence measures provide visual cues for attribute combinations of interest. Users can study multiple selected groups in the group panels (D1 and D2), where they can also compare them. In each panel, three tab views can be switched to visualize: (1) attribute measures as parallel coordinates plots (see Fig. \ref{fig:PCP}) including TP, TN, FP, FN and accuracy, precision, etc.; (2) image clusters (D1) and (D2) based on their distance in different data spaces; (3) image thumbnails (See Fig. \ref{fig_CaseRings}). All these views are coordinated for synchronized changes. Clicking any image instance also adds it to a detail image view (E). The attribute values of actual labels and predicted labels are visualized. Next, we discuss details about visual exploration with these views.

\subsection{Image Exploration in Embedded Spaces}

When a set of x-ray scattering images is loaded into the system, the corresponding structural attributes are loaded and shown in the attribute panel (Fig. \ref{fig_interface}B). In each data space (ACT, FEA, and PRD), the image vectors are plotted into a 2D space as points through t-SNE projection. Users select these points in one space and meanwhile, the images are highlighted in other spaces. By comparing their distributions, users are hinted for the model behaviors. For example, a close group of images in ACT means that the images have similar actual labels of attributes. But their distributions in PRD space may be far away indicating that the model made wrong predictions. Similarly, departing images in FEA space shows that the neural network finds different high-level features in these images. Then other views provide tools for users to further study these findings.

\begin{figure}[t]\centering
 \includegraphics[trim={0 0.5cm 0 0},clip,width=0.7\columnwidth]{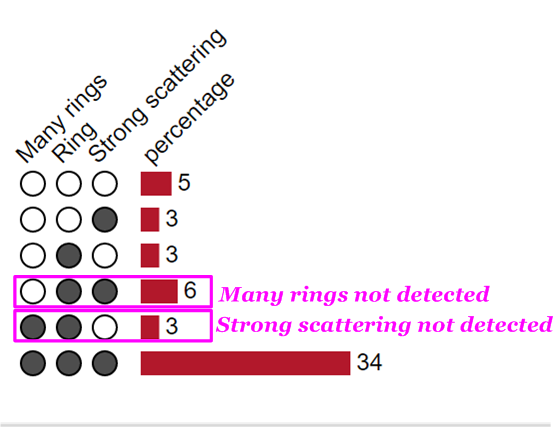}
 \vspace{-10pt}
    \caption{An attribute set view in the attribute panel showing the numbers of images with different attribute combinations. } \vspace{-15pt}
    \label{fig:attributepanel} 
\end{figure}

\subsection{Image Exploration from Attributes} \label{sec:attributes}
An alternative exploration path is to study the images from their attributes. First, users can interactively select any  combinations among all attributes in the attribute panel. Then an attribute set view shows the model performance of the images with these attributes. In Fig. \ref{fig:attributepanel}, "Many rings", "Ring" and "Strong scattering" are selected as an example. The bottom row shows there are 34 images that are correctly classified (by the dark dots). The top row indicated that all the three attributes are not correctly classified in 5 images. Users can click any row so that all the images in the corresponding set are selected for further study.




\begin{figure}[t]\centering
 \includegraphics[width=\columnwidth]{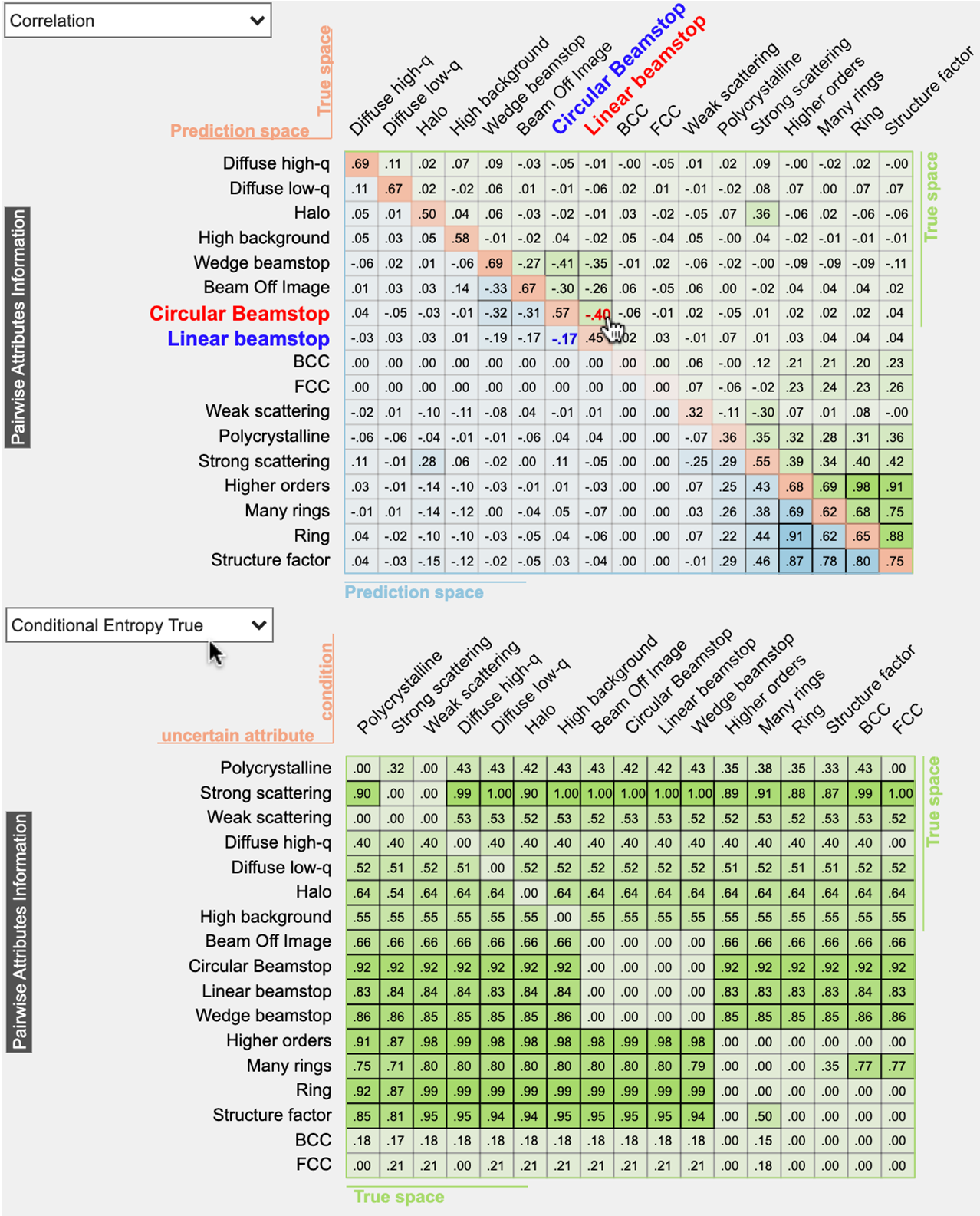}\vspace{-5pt}
    \caption{ Top: Matrix view of pairwise attribute correlation. The green triangle represents the correlation of two attributes in True space, while the blue triangle shows the correlation in Prediction space. Bottom: Matrix view of conditional entropy in True space.}\vspace{-15pt}
    \label{fig:matrix} 
\end{figure}

Users can further discover image instances based on co-existence metrics. We visualize these metrics in interactive color-enhanced matrices. For both Pearson correlation and mutual information,  the generated matrix is symmetrical since changing the order of two attributes will not affect the dependence result. To save space, we use  half of a matrix to visualize one relationship of two attributes, as shown in the top of Fig.\ref{fig:matrix} where the green and blue triangles form a full matrix. The green triangle represents the correlation of two attributes in the true space, while the blue triangle shows the correlation in the prediction space. In addition, the values in the diagonal cells are computed with one attribute from the true space and the other from the prediction space and colored in orange. A high-value cell in the matrix indicates a strong correlation. Users can identify interesting pairs of attributes in one triangle, while the same pairs in the other triangle are also highlighted. This design helps users evaluate the model performance by comparing the relationship between the two attributes in two spaces.   

For conditional entropy, the order of the two attributes matters. A full matrix is used for the attributes in either true space or prediction space. The horizontal attributes are conditional attributes, and the vertical ones are the uncertain attributes. We show the matrix for the true space at the bottom of Fig.\ref{fig:matrix}. For example, zero value cells of the matrix indicate the two attributes are either simultaneously existing (e.g., high orders and many rings) or are mutually excluded (e.g., circular beamstop and linear beamstop). 

\begin{figure}[t]\centering
 \includegraphics[trim={0 0 1cm 0},clip,width=\columnwidth]{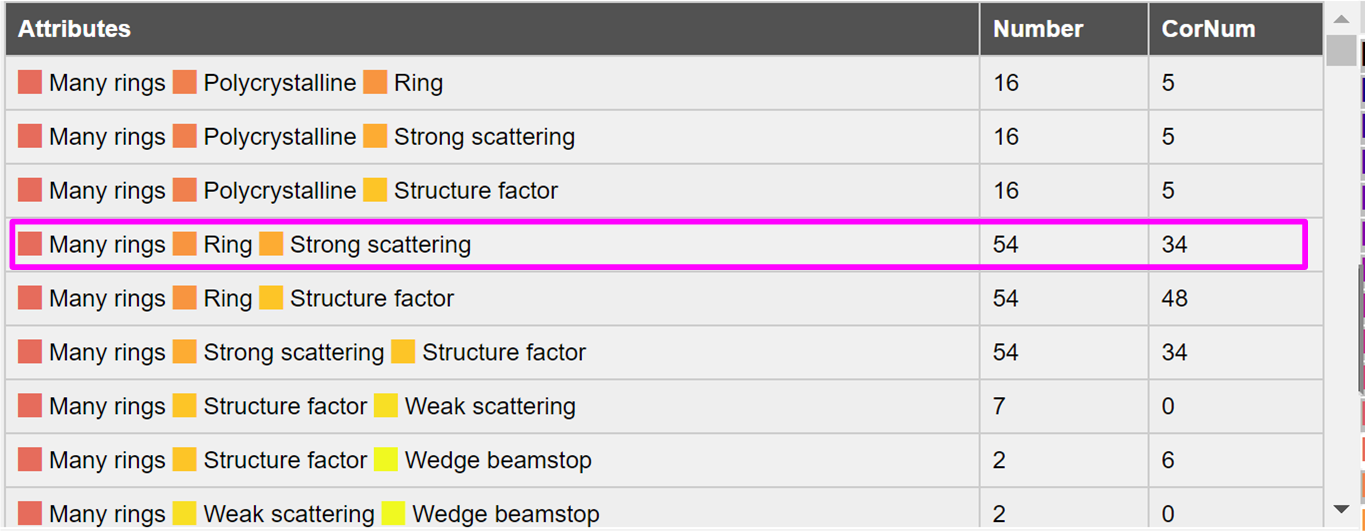}\vspace{-5pt}
    \caption{ A co-existence table view of multiple attributes. It shows the number of images having the attribute combination in ACT space (Number), and the number of correctly predicted images (CorNum).
    }\vspace{-11pt} 
    \label{fig:coexistencetable} \end{figure}

In addition, Fig. \ref{fig:coexistencetable} is the table view of multi-attributes. Here users can choose a different number of attributes and three attributes are shown in this figure. The top three attributes can be ranked (with user interaction) by the total number of images with them or the number of correctly predicted images. Users can select any group of images for further study.

\begin{figure}[t]\centering
 \includegraphics[width=\columnwidth]{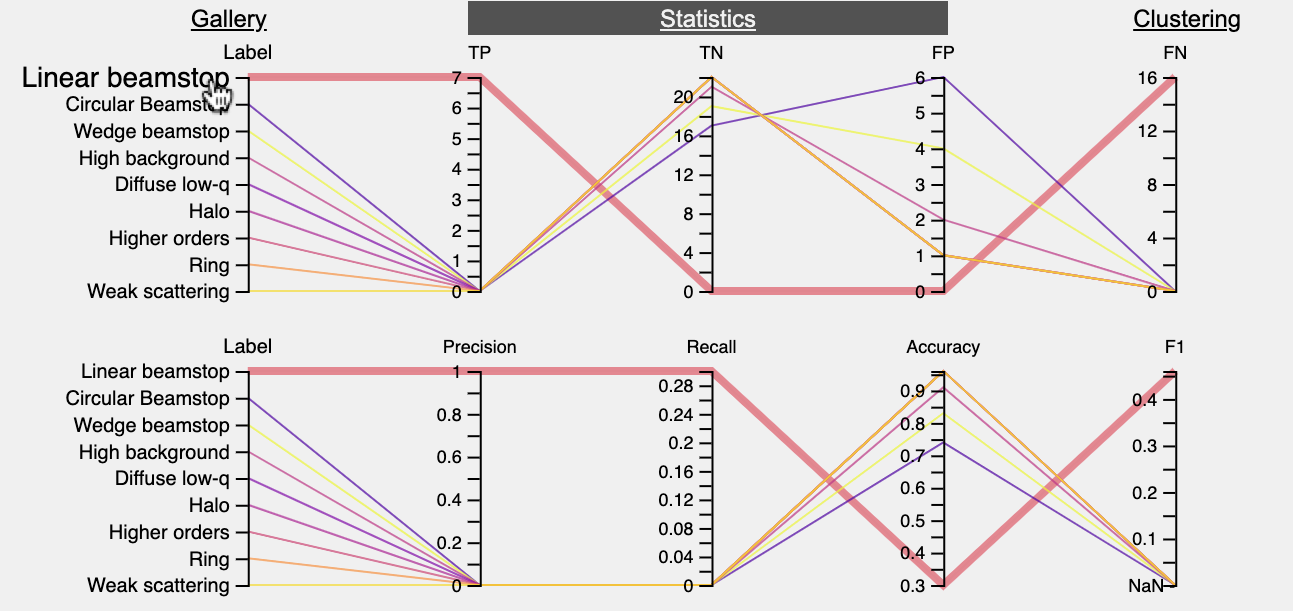}\vspace{-10pt}
    \caption{ Two parallel coordinates plots showing attribute measures of model performance with various metrics. }\vspace{-15pt}
    \label{fig:PCP} 
\end{figure}

\subsection{Image Group Exploration and Comparison}
While the groups of images are selected, they can be drilled down for the characteristic measures as shown in Fig. \ref{fig:PCP}, where two parallel coordinates plots (PCP) are used to show the measures of each attribute inside one group. On the left, the attributes that exist in most images are shown on the top. Users can hover over each attribute to highlight the polyline, while the corresponding images are highlighted in the embedded space views as well. 

\begin{figure*}[t]
 \centering
 \includegraphics[width=0.83\textwidth]{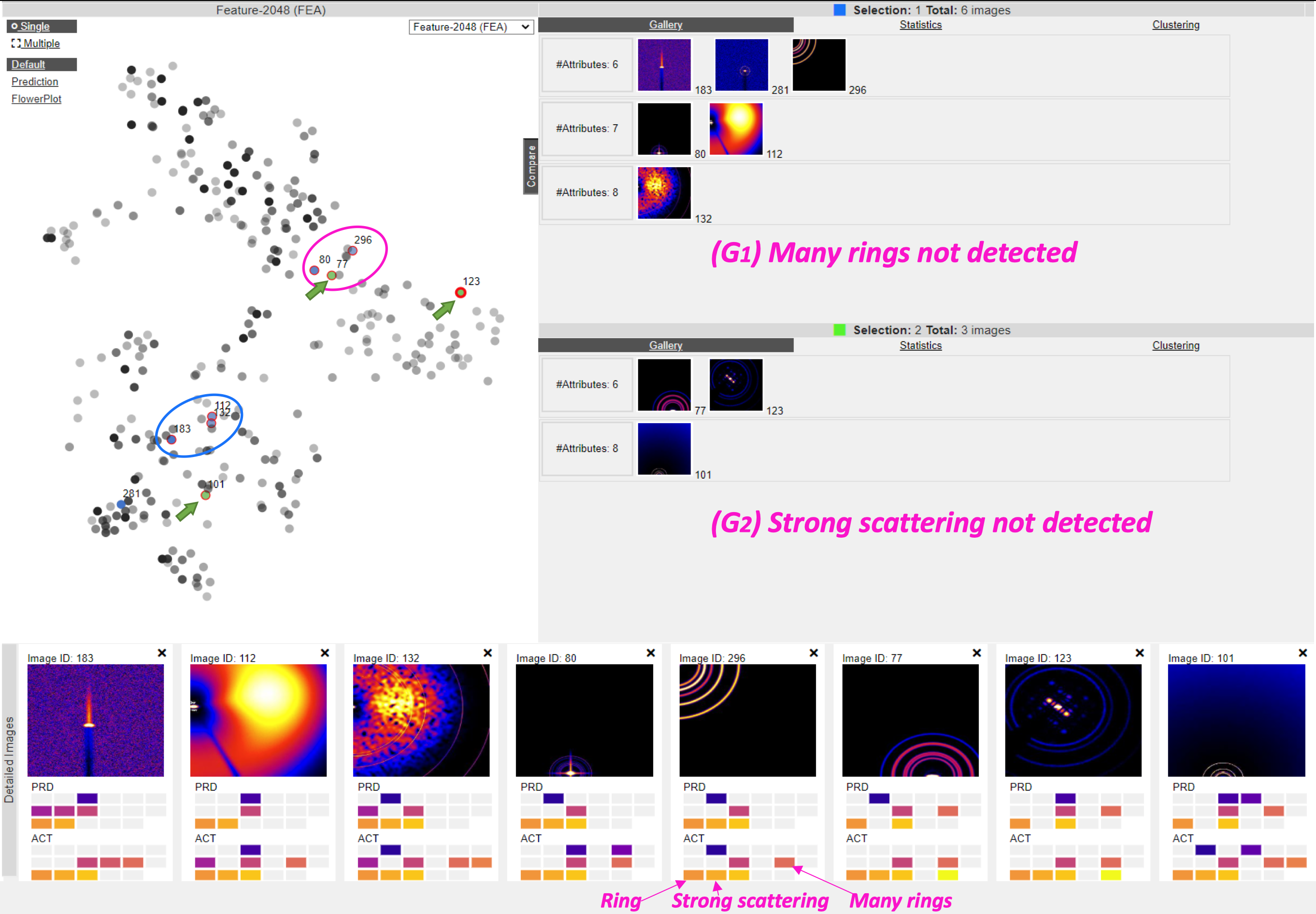}\vspace{-5pt}
    \caption{ Studying two groups of images where attributes ``Many rings'', ``Ring'', and ``Strong scattering'' co-exist. (G1) ``Many rings'' not detected; (G2) ``Strong scattering'' not detected. The image galleries and the FEA space are shown, together with the detail view of several images.}\vspace{-15pt} 
    \label{fig_CaseRings} 
\end{figure*}

The selection of images can also be visualized in a gallery with thumbnail views, as shown in Fig. \ref{fig_CaseRings}. Here, the images are grouped by the number of attributes they include (in ACT or PRD). It is easy for users to identify outliers that may be wrongly labeled of attributes.  

Moreover, these images can be displayed in a cluster view as shown in Fig. \ref{fig_interface}(D1-D2). The motivation of using the clustering methods here is to help users discover sub-groups of images so that the data distribution patterns can be studied and compared easily. There are difficulties in finding the absolute optimal solution of clustering \cite{OptimalClustering,Kleinberg2002} due to different distance metrics and user preference. Therefore, two methods, K-Means as a centroid-based clustering and DBSCAN as a density-based clustering, are provided so that they have the flexibility to interactively find good clustering results for drill-down study. DBSCAN can find clusters with respect to spatial distribution while K-Means favors globular clusters. To help them in verifying the clustering, two unsupervised clustering validation scores, Silhouette Coefficient \cite{SilhouetteScore} and Davies-Bouldin score\cite{DaviesScore}, are computed and visualized in real time. For example, Fig. \ref{fig_interface}(D1) shows two \textit{spatial clusters} generated by applying DBSCAN to the blue image points in PRD space in Fig. \ref{fig_interface}(C3). This clustered view provides a more clear depiction and easier interaction of the images. Users can control the tuning parameters, ``EPS" and ``MinPts" for DBSCAN to generate a good clustering result based on the validation sores. Similarly, in K-Means, the number of clusters can be tuned directly. Moreover, users can cluster the same selection of images according to the different latent spaces in  Fig. \ref{fig_interface}(D1 and D2).  For example, in contrast with the clusters in PRD space in Fig. \ref{fig_interface}(D1), the same selection of images highlighted by the blue image points were dragged into Fig. \ref{fig_interface}(D2) where it shows the two clusters by K-Means method in FEA space and matches with the points distribution in Fig.\ref{fig_interface}(C2). 

\begin{figure}[t] \centering
 \includegraphics[width=0.6\columnwidth]{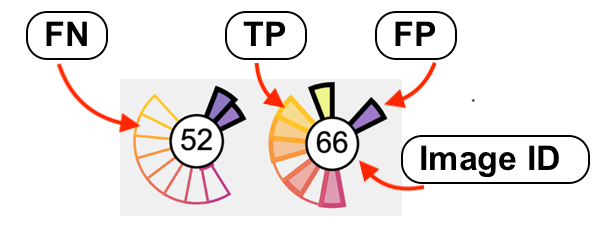}\vspace{-15pt}
   \caption{ The attribute flower visualization where each petal represents one attribute. A blank petal indicates FN, a solid petal indicates TP, a missing petal indicates TN, and a solid petal with a black border means FP. }\vspace{-15pt}
   \label{fig_flower} 
\end{figure}

In addition, an \textit{attribute flower} visualization (Fig. \ref{fig_flower}) is designed (based on Astor charts) to show each attribute as a ``petal''. Each petal is filled in the corresponding color of one selected attribute if the image has this attribute in its ACT vector. The petal has a black border if this attribute does not exist in the ACT vector but appears in its PRD vector. The missing petals simply indicate true negative prediction. Therefore, FN, FP, TP and TN attributes can be easily discerned for each image and for comparison.

\section{Case Studies}
We have conducted several case studies with two domain experts in physical and material sciences who have applied DL for x-ray data analysis and one graduate student who has worked on ResNet model development for x-ray data. They operated our Web-based system remotely through web browsers.

\subsection{Case 1: Studying model performance from an ACT image group}

Fig. \ref{fig_interface} shows an image group (in blue dots) selected from ACT space, based on its high prediction error rate (the dots with higher opacity). These images are the most problematic candidates. By putting this selection into the group panel, Fig. \ref{fig:PCP} indicates that a structural attribute ``Linear beamstop'' is the most important in this group. Switching to the clustering view in Fig. \ref{fig_interface}(D1), the images form two spatial clusters in PRD space. The red petal of the attribute flowers indicates ``Linear beamstop''. In the bottom cluster, images with this attribute are predicted correctly such as Image 54, 62, and 139. From the detail view (Fig.\ref{fig_interface}(E)), the three images are very different (Image 62 has a blue background and needle shape under the beam) while the learning model gives a correct prediction on this attribute. 

However, most images in the top cluster of Fig. \ref{fig_interface}(D1) are wrongly classified. For example, Images 44, 38, and 4 fail the detection of linear beamstop. Instead, they are detected as other types of beam stops: ``Wedge Beamstop'' or ``Circular Beamstop''. Referring to Fig. \ref{fig:matrix},  the coefficients of the correlation matrix between any pair of these three different beamstops are negative, telling they are negatively dependent and exclusive to each other in an image. The values of conditional entropy between them in the green true space are zero, which further verifies they are exclusionary. If users switch to the conditional entropy in the predication space from the drop-down box, they will find that the conditional entropy values for these pairs are non-zero. Thus, it shows the model cannot tell the differences of the three different beamstop attributes.  

By further putting this group into Fig. \ref{fig_interface}(D2) and using FEA space instead with 2 spatial clusters, Images 44, 38, 4, and 139 are close together. The convolutional part of the network finds they have similar high-level image features. But the fully connected classifier may be confused and give different predictions of beam stop types. This shows that the x-ray scattering attributes of different types of beam stops are quite hard to detect due to the small size and especially with a dark black background. Therefore, the visualization system helps scientists recognize the weak point of a learning model, which may be addressed by further extending the convolutional layers and training the model with more images to distinguish them in the feature space. 

\subsection{Case 2: Studying model behavior with three co-existence attributes}

From the 3-attribute co-existence table of Fig. \ref{fig:coexistencetable}, three attributes of interest are identified: ``Many rings'', ``Ring'', and ``Strong scattering''. These attributes are highly correlated and scientists want to confirm how the network distinguishes them. In the dataset, there are totally 54 images sharing the attributes which are correctly detected in 34 images. By selecting these attributes, from the attribute set view of Fig. \ref{fig:attributepanel}, users can identify different combinations and find the model output of them. For example, two combinations are selected to groups of interest: (G1) six images where ``Many rings'' are not detected; (G2) three images where ``Strong scattering'' are not detected. The two selections are put into the group panel for comparison, as shown in Fig. \ref{fig_CaseRings}. Here the thumbnail galleries of them are shown. Note that all the images have the attribute ``Ring'' detected correctly while their visual appearances are quite diverse, where the high variation of x-ray scattering structures are manifested. 

In the embedded view of FEA space, G1 images are shown in blue dots and G2 images are in green dots (the color can be selected by users flexibly). It can be seen that G2 images are far away from each other. But Image 77 in G2 are very close to Images 80 and 296 in G1 (see the pink circle). When opened in the detail view, the three images have similar features of rings whose center roughly lies on edges and corners. The model detects many rings in Image 77 but not in Image 296. It helps scientists to identify that the model does not perform as expected for ``Many rings'' due to the fact that the rings in Image 296 are not as complete as in Image 77. Here, since high-level FEA features are extracted correctly, to improve the model, the fully connected layers may be replaced with other classifiers (e.g., Support-Vector Machine).

In G1 images, 183, 112, and 132 are close in FEA space (see the blue circle). In the detail view (first three images), they have specular backgrounds with strong scattering (which is related with Bragg angle of light beams). The ``many rings'' structure is quite hard to discern from low-level pixel-based metrics. This type of ``Many rings'' may need to be labeled in a separate way, which suggests scientists to use a refined attribute list that sufficiently describes the image structures.

\subsection{Case 3: Studying the pre-trained model error}

\begin{figure*}[t]\centering
 \includegraphics[width=0.83\textwidth]{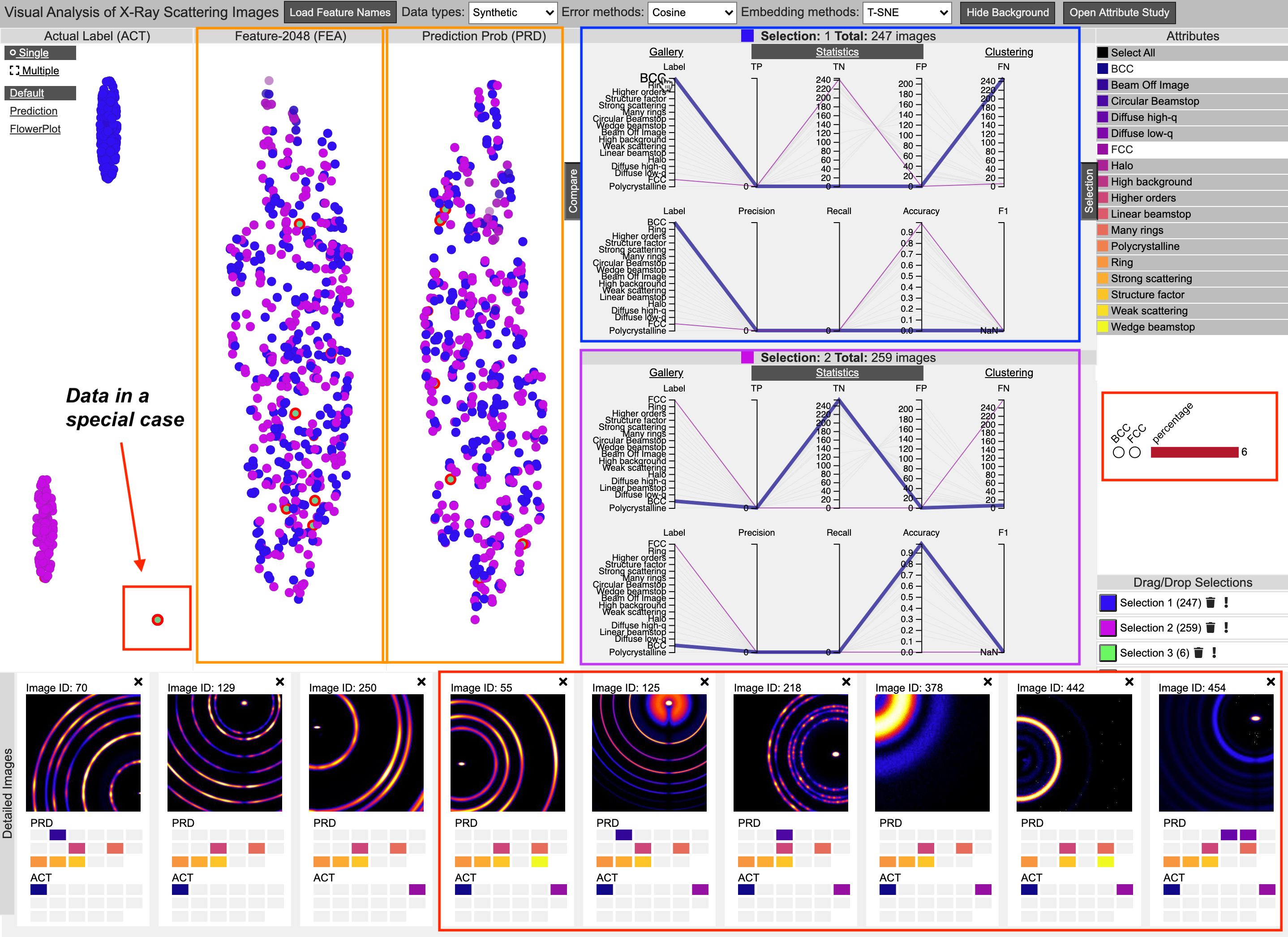}\vspace{-5pt}
    \caption{ Debugging the issue of the pre-trained ResNet model for two attributes BCC and FCC through the entire visualization elements. }\vspace{-15pt}
    \label{fig:Debug_case} 
\end{figure*}

By exploring the co-existence matrices in Fig. \ref{fig:matrix}, we notice that two attributes  ``BCC'' (body-centered cubic) and  ``FCC'' (face-centered cubic) are presented as anomalies. For example, in the correlation matrix (the top one in the figure), the values paired with ``BCC'' in the prediction space are zero, which are inconsistent with the values in the true space. The other attribute ``FCC'' is similar. Moreover, the conditional entropy in the true space (the bottom matrix in the figure) for the two attributes present normal relationships with other attributes. Both observations indicate the predictions by the pre-trained model are erroneous for these two attributes.

As shown in Fig. \ref{fig:Debug_case}, we study the two attributes by filtering out all other labels in the attributes list. In the ACT space, three clusters are formed that are highlighted in blue, purple, and green (highlighted in the orange box), correspondingly. They represent the image clusters with only ``BCC'', only ``FCC'', and both attributes, respectively. These two attributes describe the feature of crystal structures. Typically, since the atoms in a unit cell can only form one structure, the two structures are mutually exclusive to each other. But some material samples can have two structure units superimposed. The six abnormal image points, shown as a small group in the orange box of ACT space, are identified. Next, we examine the two dense clusters in blue and purple colors in ACT space to learn more details. These two attributes cannot be separated well in both FEA and PRD spaces. As shown in Fig. \ref{fig:Debug_case}, the image points with purple and blue colors are mixed in the two embedded visualizations. 

To find out the reason, we check the performance of the two groups by looking into the parallel coordinates in the group panel (in the middle of Fig. \ref{fig:Debug_case}). First, Selection 1 (the blue cluster in ACT space) shows that all 247 images in this selection with attribute  ``BCC'' have zero true positives and zero true negatives. The accuracy, precision, and recall are all zeros and the F1 score shows NaN for  ``BCC''. It means the model is unable to predict the ``BCC'' attribute. At the same time, we find the model does not confuse the  ``BCC'' attribute to  ``FCC'', since this group has all images predicted true negative for ``FCC''. So it has a good accuracy value 1 for  ``FCC'' prediction. In Selection 2, the situation for attribute  ``BCC'' and  ``FCC'' in parallel coordinate curves is the same as in Selection 1. Both of these indicate that the ResNet model does not successfully classify these two attributes. The possible reason could be: 1) insufficient data labeled with these attributes in the training set; 2) the feature of these attributes are too similar to distinguish; or 3) they are affected and overwritten by other strong signal attributes. 

To further check the reason, we click on some of the images from the two groups to observe their original x-ray images, as shown in the image view at the bottom of Fig. \ref{fig:Debug_case}. The  ``BCC'' and ``FCC'' attributes can be found by domain scientists on the colorful spots with periodical intensity variation on the rings. They are not suppressed by other attributes like "many rings". We also find that a quite small number of images with these two attributes were used in model training. Finally, we check the model implementation and identify an error in the data processing function that learned incorrect names for the two attributes. This example shows that the visualization system helps users discover mistakes or errors in modeling.
\section{Evaluation and Discussion}
The visualization system was evaluated in two stages. One is to get feedback from domain scientists about the usability, limitation, and suggestions of the system. The other is to evaluate the visual interface functions. We also discuss the system from multiple aspects.

\subsection{Interview with Domain Scientists}

We conducted an interview with the two domain scientists who have contributed to the case studies. One of the scientists is a co-author. They were given a thorough introduction to the system and its functions. Then, they practiced freely with the system for the investigation of the x-ray scattering data and performed the following study. 

\subsubsection{A Usage Scenario:}
Domain experts found two groups with images that show big ``ring'' and small ``ring'' on FEA space. In the semantic PRD space and ACT space, these two groups are supposed to indicate the same feature. However, they are separated in the FEA space, which means the neural network takes the scale of the attribute ``ring'' into account significantly. This is an interesting point found by experts. It means that in the FEA space, the clusters are formed by the data points visually close to each other rather than semantically. Thus, the function of the fully connected layers is to collect the visual features from a widely distributed area in FEA for classification. Since ResNet50 contains only a single fully connected layer to do this work, the domain experts expected that more fully connected layers may be added in the end of the classification phase.   

\begin{table}[t]				
\caption{QUIS questions and ratings }			\centering				
{\makebox{\resizebox{\columnwidth}{!}{				
\begin{tabular}{l|c|c} 
\hline
\multicolumn{0}{c|}{Questions}  & Means & Standard \\	
 &  &  Deviation\\ \hline	
 
\textit{Part I: Visual Tools Questions: 0(poor) - 9(excellent)} &\\
\rowcolor{Gainsboro!60}
\textit{Coordinated Visualization with Embedded Spaces} & &\\
Embedded views for showing image groups and outliers. &8.45 &0.52 \\			
Zooming and panning operations.&8.91 &0.30 \\				
Multiple spaces comparison and drill down study. &8.82 &0.40\\
Image item transparency showing errors. &8.27 &0.79 \\			
Image item labeling. &8.36 &0.67\\

\rowcolor{Gainsboro!60}
\textit{Image Group Selection}& &\\
Free selection on embedded spaces. &8.91 &0.30 \\	
Image group management. &8.64 &0.50 \\	
Image group coloring function.&8.82 &0.40 \\

\rowcolor{Gainsboro!60}
\textit{Image Group Visualization}& &\\
Group panel drag and drop.  &8.73 &0.64 \\
Group measures with PCP. &8.27 &1.27 \\
Group thumbnail gallery overview. &8.63 &0.50 \\
Group spatial clustering. &8.73 &0.47 \\
Image attribute flower view. &8.73 &0.64 \\

\rowcolor{Gainsboro!60}
\textit{Attribute Co-existence Visualization}& &\\
Attribute list with interaction.  &8.64 &0.50 \\
Attribute set view and group selection. &8.36 &0.57 \\
Attribute co-existence table view.  &8.63 &0.57 \\

\rowcolor{Gainsboro!60}
\textit{Detail View Image Visualization}& &\\
Image visual comparison. &8.64 &0.67 \\
Image ACT and PRD value comparison. &8.45 &0.52 \\

\hline

\textit{PartII. Visualization system rating} &\\
\rowcolor{Gainsboro!60}
\textit{Interface} & &\\	
Reading labels and icons. 0(very hard)-9(very easy)&8.10 &1.04\\				
Highlighting selected focuses. 0(not at all)-9(very much)&8.27 &0.65\\	
Information organization. 0(confusing)-9(very clear)&8.45 &0.69\\				
Sequential operations. 0(confusing)-9(very clear)&8.18 &0.98\\			
Interactions. 0(very hard)-9(very easy)&8.23 &0.79\\

\rowcolor{Gainsboro!60}
\textit{learning} & &\\
Learning to operate the system. 0(difficult)-9(easy)&7.18 &1.17\\
Performing tasks is straightforward. 0(never)-9(always)&8.09 &0.83\\

\rowcolor{Gainsboro!60}
\textit{System} & &\\				
System response speed. 0(very slow)-9(fast enough)&8.36 &0.81\\				
Designed for all levels of users. 0(never)-9(always)&7.82&1.32\\		
System reliability. 0(unreliable)-9(very reliable)&8.18 &0.75\\		
\hline
\end{tabular}				
}}} \vspace{-15pt}
\label{table:QUIS}				
\end{table}	

\subsubsection{Feedback on Usability} 
Based on the experience with the usage scenario, the domain scientists provided feedback on the system usability in research:

\noindent \textbf{Good points:}
\begin{itemize}
\item \textit{About system interface:} This interface allows users to explore the training data and the trained model side-by-side. The tool is very easy to explore and identify the issue of the model.

\item \textit{About interlinked embedded views:} The linking of data between embedded views is extremely important and well-executed. This allows users to follow a chain of logic from one representation to another to study model performance.

\item \textit{About outliers and hypotheses:} The system allows users to explore and search for outliers, as well as to test hypotheses by grouping data and performing analytic tests. This kind of interaction is extremely useful for being able to then refine the training model. 

\item \textit{About model behavior:} Perhaps, more importantly, this tool allowed users to identify limitations and biases in the training data itself. Thus, this tool is a useful guide towards improving training datasets.
\end{itemize}

\noindent \textbf{Limitation and suggestion:}
\begin{itemize}
\item \textit{About network layers:} Only the feature space after all convolutional layers is used, it would be good to further show the study of the feature maps from other layers in the deep network.

\item \textit{About embedded space control:} For embedded spaces, it is very useful to see how data is organized. 
However, because these spaces are transformed from the high dimensional spaces, it would be nice to have more control over how one is viewing these embedded spaces, such as using alternate layout modes, ability to rotate the view, etc.

\item \textit{About group study:} For group study panel, it requires some training to know how to use them and interpret the information. Perhaps some additional feedback to users could make these tools more immediately understandable.
\end{itemize}

\subsubsection{Feedback on System Impact} 
The domain scientists agree that this visualization tool provides an easy way to explore tagged datasets from the output of a trained machine-learning model. The primary advancement herein is providing accessibility of this meta-data to the domain expert. In their current workflows, the domain experts search meta-data in an \textit{ad hoc} manner using database-like interfaces. They have no ability to search or visualize the tagging outputs of machine-learning models, other than to evaluate statistical metrics such as average precision. The presented tool thus provides a way for domain experts to easily study the machine-learning annotations of their data. This allows them to build confidence in their machine-learning models, to direct improvements in those models, and (eventually) as a way to browse their data looking for interesting meta-data correlations that would be hard to otherwise see.  In a conventional experiment, the researchers collect large amounts of data in the model, then manually search these data for both expected patterns (hypothesis testing) and unexpected patterns (exploration). Once a pattern is identified, a detailed data analysis is performed in order to highlight and explain it. The proposed visualization tool dramatically alters the second step (searching data for trends) by affording the opportunity for the domain expert to identify interesting and unexpected relationships between meta-data attributes. The tool also provides a convenient interface for summarizing the results of a given experiment, and thus allows data browsing in a more convenient manner as compared to conventional workflows.

\subsection{Evaluation for Visual Interactions}


\noindent \textbf{Participants:} Eleven graduate students who are computer science majors (4 females, 7 males) participated in the evaluation of the visualization system. All the participants had  knowledge of visualization and interface design and experience in web-based visualization tools. Six of them have experiences of deep learning models and tools. 

\noindent \textbf{Procedures:} An instructor first introduced the x-ray scattering images and structural attributes, and then the visualization and interaction functions to each participant. They were guided to freely explore the system with image data for about 10 minutes. Then, they completed the tasks: (a) find outliers in an image group; (b) discover the model behavior of attributes in the image groups. Then, they provided the evaluation by filling a QUIS (Questionnaire for User Interaction Satisfaction) form \cite{Chin1988}. Finally, they gave comments and suggestions for the system.

\noindent \textbf{QUIS evaluations:} 
Table \ref{table:QUIS} shows the questions and ratings in two parts about 1) The visualization design and functions, and 2) The system performance. The mean and standard errors of the user ratings are displayed as well. The average score of the visualization and interaction functions are very good above 8.0, which indicates the users were satisfied overall. For example, the interactive operations on the embedded spaces like ``Zooming and Panning'' and ``Free selection'' received excellent rating with an average of about 8.91. ``Multiple spaces comparison and drill down study'' also got a high average score, 8.82. It indicated that the interaction in exploring data in the embedded spaces was well implemented. The ratings of the visualization system are also very good. The Web-based system performance with fast response impressed the users. They also felt the system layout and organization was easy to follow. The average rating scores for ``Learning to operate the system'' and ``Designed for all levels of users'' were good but relatively lower at 7.18 and 7.82. The standard deviations of these two were also high at 1.17 and 1.32. This is reasonable as the users need to understand the deep learning model performance over the scientific images. In future work, we will add more labels and guidance in the system to shorten the learning curve. 

The users' comments also identified system limitations to be further addressed. For example, visual clutters appear when the image labels of a large group of selected images are shown in the embedded spaces. Here, a smart labeling algorithm is needed. In addition, the selected groups are fixed and more flexibility may be added to insert and delete specific images to/from a group. Moreover, the color selection for different attributes should be adjustable by user-definition. We will further improve the system according to these suggestions.

\subsection{Discussion}
Based on our design, implementation, and evaluation of the visualization system, we discuss our approach in several aspects.


\noindent \textbf{Complexity and Scalability}
The visualization system works on about 1,000 images in the study. The computational complexity is mainly determined by the t-SNE algorithm that projects them from high dimensions to 2D, which however only needs to be performed once when the dataset is loaded into the system. The clustering methods are only applied to selected groups of images for an interactive study so that the computational performance is not a problem for real time responses. Visualizing thousands of data points and the selected clusters can be executed very fast, so those smooth interactions are easily supported. Therefore, from the computational aspect, we expect the system can be scaled to a larger set of thousands of images easily. The visual cluttering issue may arise when more points are injected into the canvas. The system supports zooming and scaling for users to investigate data points. However, the capability of effective exploration may be hindered due to the increasing scale of loaded images.

\noindent \textbf{Transferability}
Our system is built up on the input of FEA, PRD, and ACT vectors for multiple attributes image datasets. Therefore, upon the availability of these vector representations of data items from a trained model, our approach can be transferred to other contexts or settings.
Moreover, the system has also been extended to natural image datasets (e.g., CIFAR10\cite{cifar10}) with single attribute classification. It can help diagnose the reasons for wrongly predicted images.

\noindent \textbf{Limitation} \label{sec:limitation}
Overlapped transparent dots (Sec. \ref{sec:design}) may form misleading error rate. When users select them and study the details they can find the facts. Interactive lens and/or jittering tools may help solve the visual cluttering issue. In addition, the direct use of existing scientific categories of attributes in deep learning models may not be very effective. For example, ``ring'' may be further divided into different types as found in the case study. Second, the training data based on the labeling of domain scientists often suffer from more mislabels and errors than natural images. An effective labeling tool may be developed to improve the accuracy and effectiveness of the labeling process itself. Moreover, the system is built up on a trained model and datasets. It may be integrated into the training process. It will also be extended to study new incoming images. 
\section{Conclusion and Future Work}
We present a visualization system for understanding the learning model of x-ray scattering images with multiple attributes. The system allows users to visually discover the embedded distributions of feature vectors, predictions, and actual labels of these images. User interactions are supported to compare selected image instances and study their prediction results related to the attributes. In future work, we will address the limitations and extend the work for model debugging and refinement, by taking neurons and different network layers into account.

\acknowledgments{

The authors wish to thank the anonymous reviewers. 
This work was partly supported by KSU graduate assistantship, BNL LDRD grant 18-009 and ECP CODAR project 17-SC-20-SC.

}

\section*{Appendix}

All 17 X-ray image attributes and examples in Image 1-10 of Fig. \ref{fig:resnet} (A).
{
\begin{itemize} 

 \item BCC (body-centered cubic): a material structure based on the precise positioning of successive rings. It appears in Image 9.

\item Beam off image: The direct beam (point around which rings and other features are centered) is not within the image itself. In Image 5, the beam is barely on the left edge of the image.

\item Circular beamstop: A circular beamstop is used to block the direct x-ray beam, leading to a circular shadow around the direct beam. Images 4, 9 have this attribute.

\item Diffuse low-q: Broad, isotropic intensity concentrated towards the direct beam position. It appears in Image 6, 9.

\item Diffuse high-q:  Broad, isotropic intensity appearing far from the direct beam (towards edges of an image). It appears in Image 7.

\item FCC (face-centered cubic): a material structure based on the precise positioning of successive rings. It appears in Image 5.

\item Halo: A broad and diffuse ring of intensity centered on the direct beam position. It appears in Image 2, 3, 4, 6, 7, 10.

\item High background: High intensity of background signal throughout the entire image. It appears in Image 4, 9.

\item Higher order:  Multiple successive rings or halos that appear as "repeats" of a central ring but with a larger radius. It appears in Image 3, 5, 6, 8, 9, 10.

\item Linear beamstop: A straight beamstop is used to block the direct x-ray beam, leading to a rectangular shadow extending away from the direct beam. Images 2 and 6 have this attribute.

\item Many rings: The occurrence of multiple rings centered on the direct beam, as seen in Images 3, 5, 7, 8, 9, 10.

\item Polycrystalline: Material description for materials containing many distinct crystalline domains, leading to a scattering ring that is textured and broken up into multiple dots.  It appears in images 6 and 8.

\item Ring: The occurrence of at least one sharp ring centered on the direct beam, as in images 5, 6, 7, 8, 9, 10.

\item Strong scattering: Overall high intensity to scattering (high signal-to-noise). It appears in Image 2, 3, 6, 7, 8, 10.

\item Structure factor: Multiple well-defined sharp rings that are likely indicative of a well-ordered structure. It appears in Image 5, 6, 7, 8, 9, 10.

\item Weak scattering: Overall low intensity to the scattering (low signal-to-noise), making the pattern difficult to discern. It appears in Image 1, 4.

\item Wedge beamstop: A wedge-shaped beamstop is used to block the direct x-ray beam.  It appears in Image 3, 10. 
    
\end{itemize}
}

\bibliographystyle{abbrv-doi}

\bibliography{template}
\end{document}